\documentclass[conference]{IEEEtran}
\IEEEoverridecommandlockouts
\usepackage{cite}
\usepackage{amsmath,amssymb,amsfonts}
\usepackage{algorithmic}
\usepackage{graphicx}
\usepackage{textcomp}
\usepackage{xcolor}

\usepackage{color}

\usepackage{caption}
\usepackage{subcaption}
\usepackage{xcolor}

\usepackage{pgfplots}
\pgfplotsset{compat=newest}

\usetikzlibrary{
	pgfplots.statistics,
}

\usetikzlibrary{plotmarks}
\usetikzlibrary{arrows.meta}
\usepgfplotslibrary{patchplots}
\usepackage{grffile}
\usepackage{tikz}
\usepackage{tikzscale}
\usepackage{placeins}
\usepackage{dblfloatfix}
\usetikzlibrary{external}
\usepackage{array}
\usepackage{booktabs}
\usepackage{multirow}
\usepackage[]{algorithm2e}
\usepackage{csquotes}
\usepgfplotslibrary{statistics}

\usepackage{algorithm2e}

\newcommand{\play}[1]{^{(#1)}}

\newcommand{\eg}{e.\,g.~}

\def\BibTeX{{\rm B\kern-.05em{\sc i\kern-.025em b}\kern-.08em
		T\kern-.1667em\lower.7ex\hbox{E}\kern-.125emX}}

\usepackage{multicol}
\usepackage{fancyhdr}
\fancypagestyle{firstpage}{
	\fancyhf{}
	
	\fancyhead[C]{\vspace{-5mm} \small{This work has been submitted to the IEEE for possible publication. \\ 
			Copyright may be transferred without notice, after which this version may no longer be accessible.}}
	
}

\begin{document}
	\pagenumbering{gobble} 
	
	\title{Intention-Aware Decision-Making for Mixed Intersection Scenarios \thanks{This work was supported by the Federal Ministry for Economic Affairs and Climate Action, in the New Vehicle and System Technologies research initiative with Project number 19A21008D.}
	}
	\author{\IEEEauthorblockN{Balint Varga, Dongxu Yang, S\"oren Hohmann} 
	\IEEEauthorblockA{\textit{Institute of Control Systems, Karlsruhe Institute of Technology} \\
			Karlsruhe, Germany \\
			balint.varga2@kit.edu}
	}
	
	\maketitle
	\thispagestyle{firstpage}
	\pagestyle{empty}

	\begin{abstract}
	This paper presents a white-box intention-aware decision-making for the handling of interactions between a pedestrian and an automated vehicle (AV) in an unsignalized street crossing scenario. Moreover, a design framework has been developed, which enables automated parameterization of the decision-making. This decision-making is designed in such a manner that it can understand pedestrians in urban traffic and can react accordingly to their intentions. That way, a human-like response to the actions of the pedestrian is ensured, leading to a higher acceptance of AVs. The core notion of this paper is that the intention prediction of the pedestrian to cross the street and decision-making are divided into two subsystems. 
	On the one hand, the intention detection is a data-driven, black-box model. Thus, it can model the complex behavior of the pedestrians. On the other hand, the decision-making is a white-box model to ensure traceability and to enable a rapid verification and validation of AVs. This white-box decision-making provides human-like behavior and a guaranteed prevention of deadlocks. An additional benefit is that the proposed decision-making requires low computational resources only enabling real world usage.  
	The automated parameterization uses a particle swarm optimization and compares two different models of the pedestrian: The social force model and the Markov decision process model. Consequently, a rapid design of the decision-making is possible and different pedestrian behaviors can be taken into account. The results reinforce the applicability of the proposed intention-aware decision-making.

	\end{abstract}
	
	\vspace*{2mm}
	
	\section{Introduction}
	
	With the increased number of highly automated and autonomous vehicles (AVs) on the streets, the safety of vulnerable road users (VRUs, e.g.~cyclists and pedestrians) is getting more in the focus of the original equipment vehicle manufactures \cite{2018_WhatDrivesPeople_xu, 2019_PublicAcceptanceFully_liu, 2019_RolesInitialTrust_zhang}. The most challenging situations arise at low speed in urban traffic interacting with VRUs and other human-driven vehicles. In order to ensure a wide social acceptance of AVs and make the traffic safer for all traffic participants, the \textit{trust} of VRUs to AVs has to be taken into account \cite{2012_TrustSmartSystems_verberne, 2015_InvestigatingImportanceTrust_choi, 2019_LookWhoTalking_du}.
	
	\textit{Trust} means in this context that the AV and the VRU can understand each other and communicate efficiently similarly to a human-driven vehicle, see \eg \cite{2017_PedestriandriverCommunicationDecision_sucha,2020_AutonomousVehiclesThat_rasouli, 2020_LookingAheadAnticipating_chaabane}.
	Its importance is explained through the scenario illustrated in Fig. \ref{fig:scenario_1}: The pedestrian intents to cross the street, but they do not have right-of-way. Moreover, the vehicle travels at a low speed such that it could stop and let the pedestrian cross the street first, which is a typical scenario for urban traffic. Thus, for the acceptance of AVs and for the safety of VRUs, a communication between them is inevitable. Due to the fact that the presence of AVs is novel for VRUs, the communication can be problematic and dangerous. Indeed, human drivers, pedestrians and cyclists communicate with hand signs and eye contact, which is a learned way to increase the trust of VRUs~\cite{2015_PedestrianStareDrivers_gueguen, 2018_UnderstandingPedestrianBehavior_rasoulia}.
	
Due to the absence of a human driver in an AV, communication with eye contacts and gestures is not possible. Furthermore, there has only been a focus on explicit or implicit communications between AVs and VRUs in the recent years \cite{2018_WhenShouldChicken_fox, 2020_DefiningInteractionsConceptual_markkula, 2022_ContinuousGameTheory_camara}. Such an \textit{intention-aware decision-making}\footnote{Note that in this paper, the term \textit{controller} is used as a synonym for the \textit{intention-aware decision-making} of a vehicle.} of an AV has three main elements:

	\begin{figure}[t!]
		\centering
		\includegraphics[width=0.999\linewidth]{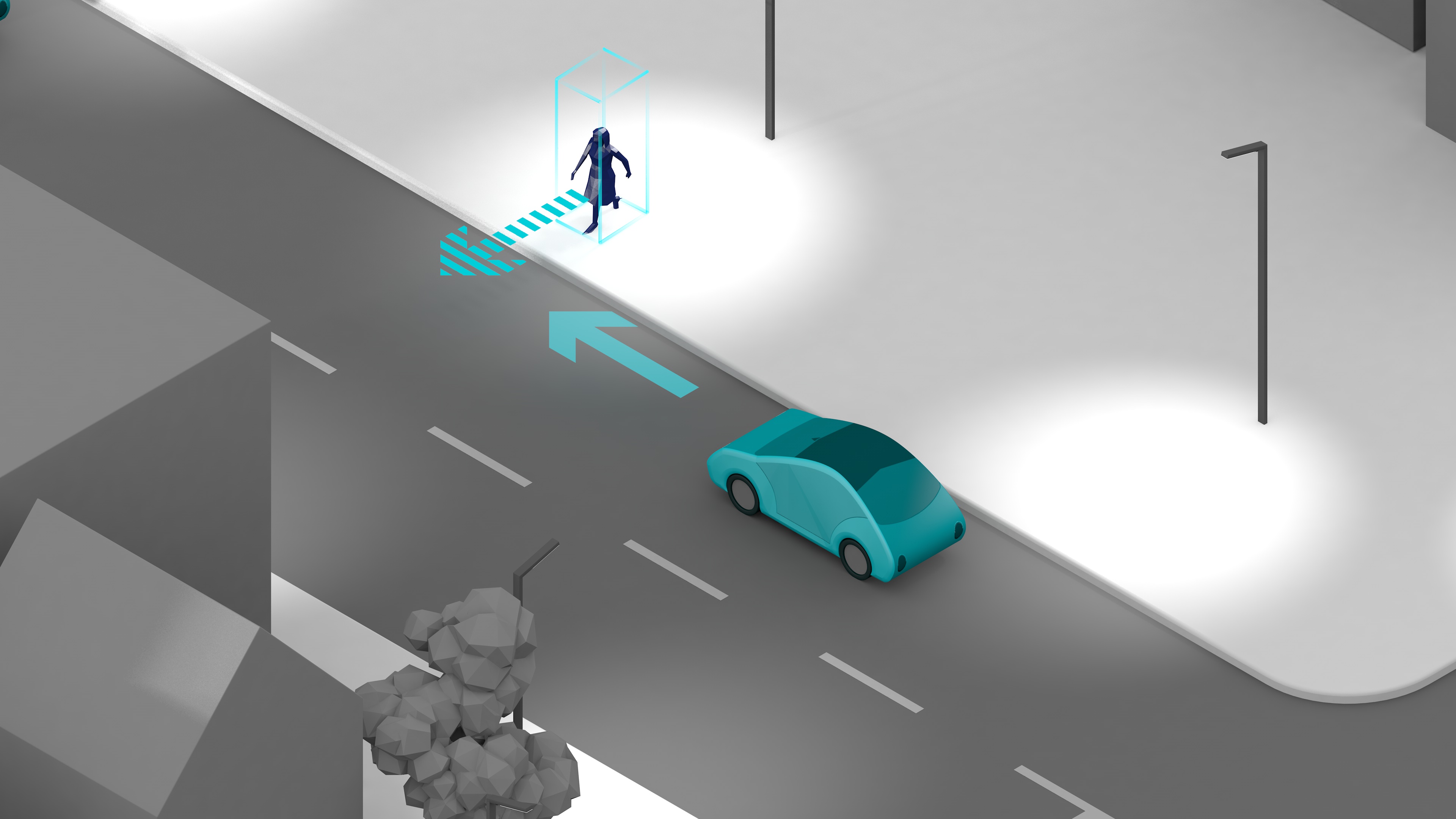}
		\caption{The pedestrian is crossing the street leading to a mixed intersection scenario, which necessitates the communication between AV and VRU in order to ensure the safety of the pedestrian. With courtesy of version1 GmbH.}
		\label{fig:scenario_1}
	\end{figure}
	\begin{itemize}
		\item[1)] Understanding the intentions of the pedestrian.
		\item[2)] Making a decision, which ensures safety of the VRUs.
		\item[3)] Communicating the decision with the VRUs.
	\end{itemize}
	In order to provide a high acceptance, all these three steps have to imitate a human-like behavior.	
	Fig. \ref{fig:software_struct_1} illustrates the software and hardware structure of the proposed technical system including these three aforementioned elements. 
	
	The first element, interpreting the intention of the VRU, is managed by the sensor systems (stationary\footnote{It is assumed that there are stationary/static infrastructure in the near future, which can provide additional information for AVs. There are also specific test fields, which are utilizing and testing such {stationary/static~infrastructure~\cite{2020_TrafficSensorData_zipfl}}.} and vehicle sensors) and the intention estimation. The intent estimate needs to be able to characterize the complex behavior of pedestrians, which impedes its traceability and easy verification. Furthermore, pedestrian intention estimation and prediction are well studied in literature, see \eg \cite{2012_DevelopmentPedestrianBehavior_tamura, 2020_EfficientBehaviorawareControl_jayaraman, 2022_IntendWaitCrossModelingRealistic_rasouli}. This intention estimation block provides the information for the intention-aware decision-making algorithm of the AV, whether the pedestrian plans to cross the street. Finally, the decision has to be communicated to the VRUs, which happens with an external human machine interface (eHMI), see \cite{2021_LichtbasierteKommunikationsschnittstelleZwischen_baumann} for more details.
	
	The focus of this contribution is the decision-making, which raises the questions: How can a human-like and safe decision process be realized for the aforementioned scenario needing little computing capacity and enabling real-time implementation? How should the controller be designed to enable transferability to different requirements? In order to answer these questions, this paper proposes a simple white box controller enabling human-like decision-making of the AV. 
Even though the required computing capacity is important for real-world application, it is not properly addressed in literature. 

The paper is structured as follows: Section \ref{sec:soa} presents the state-of-the-art solutions for intention-aware AVs. 
In Section \ref{sec:control_setup}, the algorithm of the intention-aware AV is presented. In Section \ref{sec:control_setup}, the design framework is introduced followed by the presentation of the results in Section \ref{sec:results}. Finally, Section \ref{sec:summary} summarizes the paper and provides a short outlook for further research.

	\vspace*{-1mm}
	\section{State of the Art} \label{sec:soa}
	In literature, there are various approaches to model and control interactions between VRUs and AVs. In the following, two aspects are taken into account: 1)~The prediction models of a pedestrian with and without the interaction with other road users and 2) the interacting control concepts of AVs, which can handle the interaction with VRUs.
	
	\vspace*{-1mm}
	
	\subsection{Pedestrian Models} \label{sec:ped_mod}
	\textit{Jayaraman et al.} \cite{2021_MultimodalHybridPedestrian_jayaraman} proposed a method which uses \textit{support vector machine} to calculate the probability of the gap acceptance to infer the intention of the human. Four states are used to describe the current movement of pedestrian and every state has its specific motion function. Further studies utilized the social force model (SFM) to simulate the human behavior \cite{2012_DevelopmentPedestrianBehavior_tamura,2015_WaitingPedestriansSocial_johansson,2018_SocialForceModels_chen}. The motion of pedestrian is influenced by his goal, other pedestrians and obstacles. 

	\begin{figure}[t!]
	\centering
		\includegraphics[width=\linewidth]{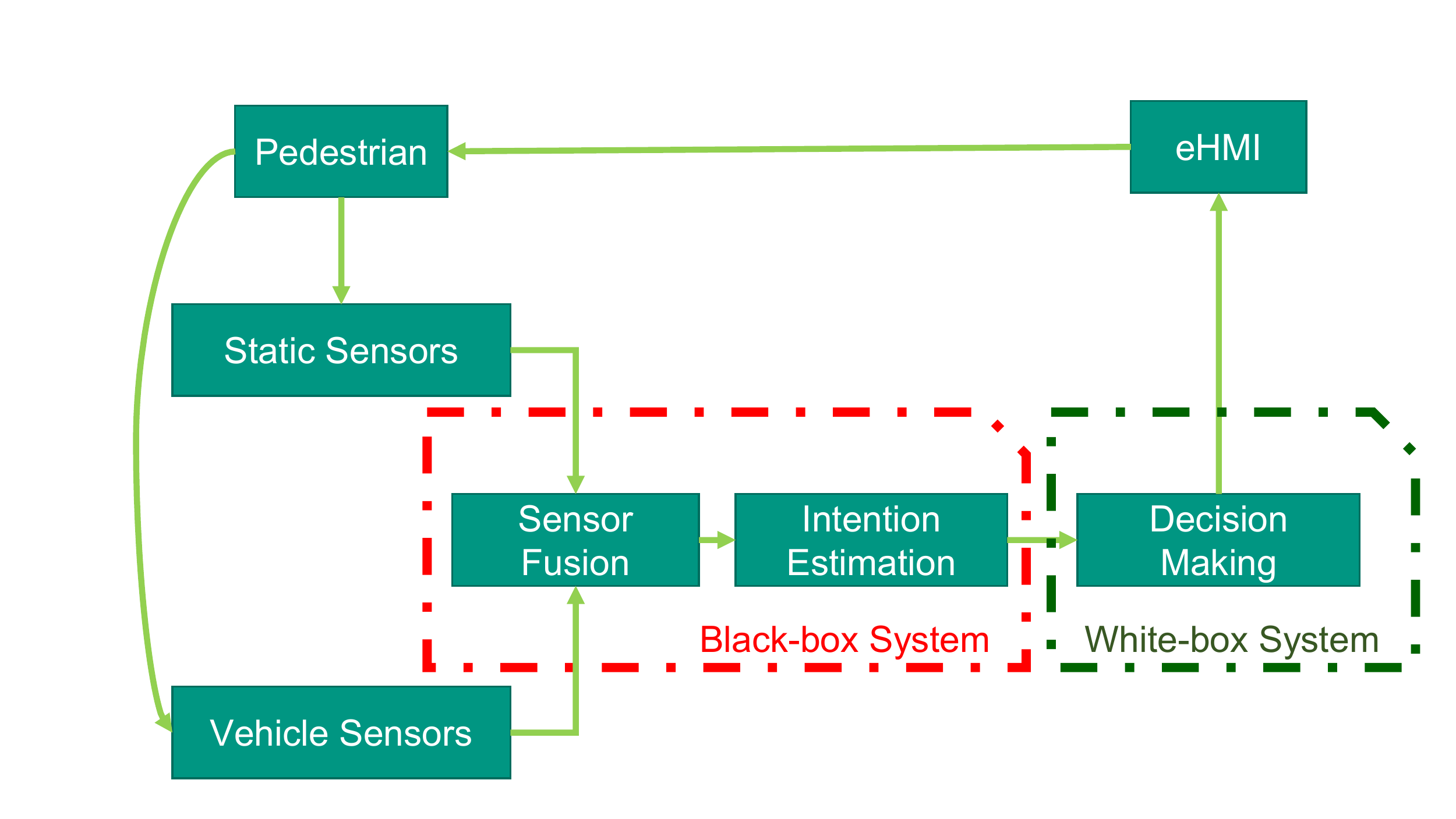}
		\caption{The proposed software/hardware structure of proposed technical system is illustrated. The term black-box implies systems with data-driven methods without traceable model structure. On the other hand, white-box denotes subsystem with a simple structure which is easily verifiable.}
		\label{fig:software_struct_1}
	\end{figure}
	
	The so-called \textit{Markov decision process} (MDP) is also a popular model for predicting the trajectory of human \cite{2015_IntentionawareOnlinePOMDP_bai, 2016_IntentawareLongtermPrediction_karasev, 2018_MDPModelVehiclePedestrian_hsu}. In \cite{2016_IntentawareLongtermPrediction_karasev}, MDP is applied to express the human behavior at signed intersection. That study also considers the influence of environment like traffic lights. Moreover, the orientation is included in pedestrian states. However, it neglects that how other traffic participants could make an effect on the movement of the human.
	In \cite{2021_LongTermPedestrianTrajectory_huang}, a \textit{long short-term memory} framework is presented that can model long-term pedestrian behavior. Due to the memory character of long short-term memory, the past information is also utilized to predict the motion in the future.
	\textit{Kooij et al.} proposed a novel method called \textit{switching linear dynamical system} to learn crucial dynamics that characterize the pedestrian near to AVs \cite{2014_AnalysisPedestrianDynamics_kooij}.

	\subsection{Interacting Control Concepts} 
	Since the behavior of VRUs can be strongly influenced by AVs during crossing, the characterization of such a interaction raises further challenges and therefore, it is crucial for an intention-aware AV. In order to solve these challenges, some studies formulated such interaction scenarios as an optimization problem, which can be solved using a model predictive controller \cite{2020_EfficientBehaviorawareControl_jayaraman}: The motion model predicts the pedestrian behavior according to a hybrid system with a gap acceptance model that only required pedestrian's position and velocity. Then a behavior-aware model predictive controller is utilized to solve the real-time motion planning problem.
	
	To model interaction between different agents in case of general problems, the state of the art utilizes game theory, see \eg \cite{2008_MultiagentSystemsAlgorithmic_shoham}. Thus, game theory is also applied for the modeling and the control of interactions between AVs and VRUs \cite{2018_WhenShouldChicken_fox, 2021_GameTheoreticApproach_amini, 2019_GameTheoryModeling_wu}.
	For instance, in \cite{2018_WhenShouldChicken_fox}, \textit{Fox et al.} proposed a negotiation model which uses a discrete sequential game for the characterization of the problem. 
	
	In \cite{2019_HybridControlDesign_kapania}, a rule-based algorithm is proposed, which uses the speed and position of the pedestrian to determine the decision of the AV. However, no further information is included, which would enable the modeling of complex scenarios. The use of a partially observable MDP (POMDP) is an effective method to simulate the uncertainties of the overall decision process of automated vehicle, see \eg \cite{2013_IntentionAwareMotionPlanning_bandyopadhyay,2015_IntentionawareOnlinePOMDP_bai, 2018_MDPModelVehiclePedestrian_hsu}. 
	\textit{Bai et al.} \cite{2015_IntentionawareOnlinePOMDP_bai} present a two-level POMDP-based planning for autonomous driving. The belief tracker outputs the probability distribution of discrete human intention as goal position from the observed pedestrian movements, and designed planner calculates the steering angle and acceleration for the agents according to the intention information. In \cite{2020_BehavioralDecisionmakingUrban_deshpande}, the POMDP model is described as Deep Q-Networks (DQN) and it combines DQN and LSTM to make decision when facing interaction with the pedestrian who wants to cross. 
	
		
	\subsection{Discussion on the state-of-the-art Solutions}
	The aforementioned models and control concepts are able to accurately simulate the movement of the pedestrian and the interactions between pedestrians and AVs. The use of complex characterization through POMDP and deep learning methods leads to more accurate models \cite{2016_PedestrianBehaviorUnderstanding_yi, 2022_PedestrianvehicleInteractionSeverity_govinda}. However, these models are not traceable and verifiable, which hinders the real world usage and the approval of automated street vehicles. Since, in \cite{2019_HeuristicModelPedestrian_camara, 2019_NegotiationVehiclesPedestrians_guptaa} it has been shown that heuristic models can model a VRU-AV interaction sufficiently good, this paper uses a simple negotiation algorithm and provides a modular structure leading to a verifiable and traceable human-like intention-aware decision-making. To this end, in the next section, the algorithm of the intention-aware decision-making is presented.

	\section{Intention-Aware Decision-Making} \label{sec:control_setup}
	
	In order to enable an interaction with the VRU, an intention-aware decision-making of the AV is necessary. Its inputs are obtained from the data-driven intention estimation block, see Fig. \ref{fig:software_struct_1}. In our current setup, a deep neuronal network is used to estimate the intention, the position, and the velocity of the VRU. Due to the modular structure, it is possible to replace this system component by other data-driven or model-based methods. 	
	
\begin{figure}[t!]
		\centering
		\includegraphics[width=0.88\linewidth]{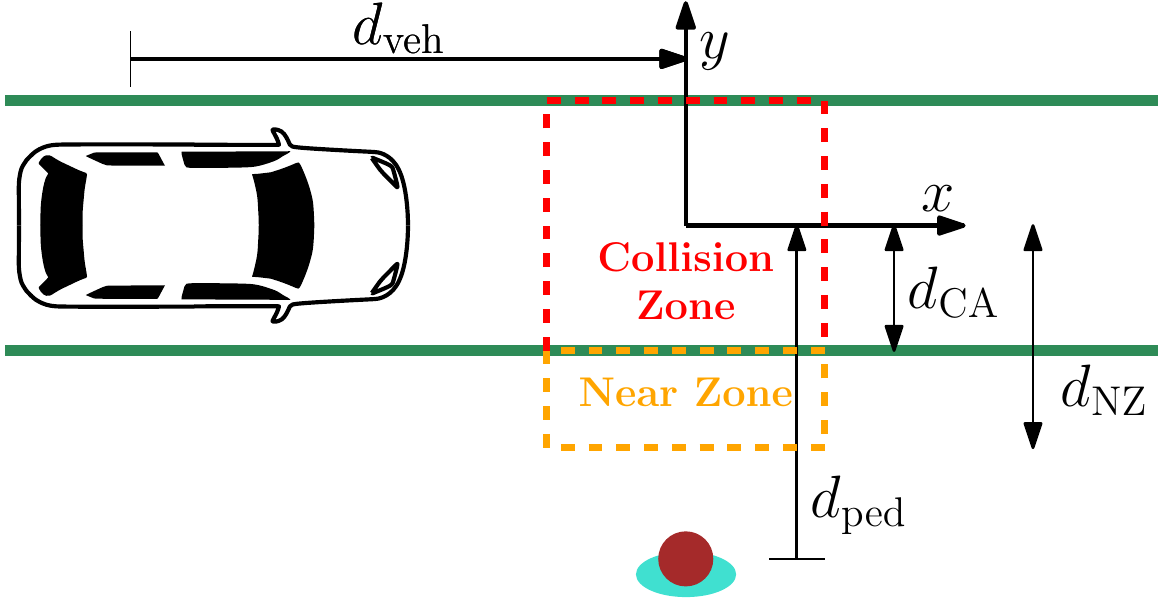}
		\caption{The scenario from bird's-eye view is illustrated, where the origin of the coordinate system is the middle of the intersection.}
		\label{fig:scenario_overview}
	\end{figure}

	The algorithm of the intention-aware decision-making is given in Algorithm \ref{alg:IAA}. The inputs of the algorithm are: The velocity $v_\mathrm{ped}$, the position $d_\mathrm{ped}$ form the intersection and the intention $i_\mathrm{ped}$ of the pedestrian, see Fig.~\ref{fig:scenario_overview}. Furthermore, the velocity $v_\mathrm{veh}$ and the position $d_\mathrm{veh}$ from the intersection of the vehicle are used. Algorithm \ref{alg:IAA} provides the necessary acceleration of the vehicle $a_\mathrm{veh,des}$ to avoid a collision with the pedestrian or to reach the desired velocity $v_\mathrm{veh,d}$ if it can cross the intersection.
	
	The structure and the parameters are kept simple in order to maintain the traceability. Furthermore, the algorithm of the intention-aware decision-making includes considerations which lead to a human-like behavior. Note that this simplicity does not restrict the usability of the algorithm, since the intention estimation can handle complex scenarios and gestures. Thus, the intention value of the pedestrian encapsulates and abstracts the potential complexity
	
	For the algorithm, the following areas are defined:
	\begin{itemize}
		\item 	A \textit{near zone} is defined by the distance $d_\mathrm{NZ}$, such that
		\begin{equation}
		\hspace*{-10mm} \mathrm{isPedestrianCloseToRoad} = \begin{cases}
				1 &  \! \mathrm{if} \; \left|d_\mathrm{ped}\right|<d_\mathrm{NZ} \\
				0 &  \! \mathrm{else}.
			\end{cases}
		\end{equation}
	\item 	A \textit{collision area} is defined using $d_\mathrm{CA}$, such that
	\begin{equation}
		\hspace*{-10mm} \mathrm{isPedestrianInCollisionArea} = \begin{cases}
			1 &  \! \mathrm{if} \; \left|d_\mathrm{ped}\right|<d_\mathrm{CA} \\
			0 &  \! \mathrm{else}.
		\end{cases}
	\end{equation}
	\end{itemize}
	
\begin{figure}[t!]
		\centering
		\includegraphics[width=0.88\linewidth]{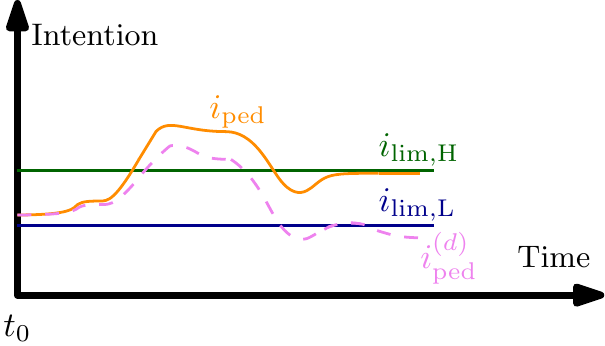}
		\caption{Illustrative representation of the decreasing the intention of the pedestrian is given, in which a high and low limit are given for the decision algorithm. The yellow line is actual intention of the pedestrian, while the purple line is the decreased. }
		\label{fig:intention_discounting}
	\end{figure}

	\vspace*{-1mm}
	The estimated pedestrian intention varies between $0$ and $1$, with the meaning $0$ - no crossing intention, $1$ - no crossing intention. In order to avoid deadlocks, continuous decreasing of the pedestrian's intention is implemented. That way, after a certain time, the AV enters the collision area and drive through even if the pedestrian had a higher intention to cross the street. Such a situation can arise when the intention estimation yield a false positive results and the AV stops even though the pedestrian does not intend to cross the street. The intention of the pedestrian is decreased such as
\begin{equation} \label{eq:discount_intention}
	i_\mathrm{ped}(t) = i_\mathrm{ped}(t_0) \cdot 0.9^{k_\mathrm{disc}\cdot t},
\end{equation}
where $k_\mathrm{disc}$ is a design parameter and $t_0$ is the beginning of the interaction between AV and pedestrian. Using \eqref{eq:discount_intention}, after a certain time, the vehicle carefully drives off. Fig. \ref{fig:intention_discounting} illustrates the proposed solution: The threshold values $i_\mathrm{lim,L}$, $i_\mathrm{lim,H}$, the lowered and the original intention $i_\mathrm{ped}$ are given. The waiting time is set through the parameters $k_\mathrm{disc}$, $i_\mathrm{lim,L}$, and $i_\mathrm{lim,H}$.

If the vehicle can cross the intersection, it accelerates to reach the desired velocity $v_\mathrm{veh,d}$, such that
\begin{equation}
	a_\mathrm{veh} = k_\mathrm{veh,acc} \left(v_\mathrm{veh,d} - v_\mathrm{veh}\right),
\end{equation}
where $k_\mathrm{veh,acc}$ the feedback gain is a design parameter.	Similarly, if the vehicle stops before the intersection, it decelerates to $v_\mathrm{veh,d} = 0\tfrac{\mathrm{m}}{\mathrm{s}}$, such that
\begin{equation}
		a_\mathrm{veh} = k_\mathrm{veh,dec} \left(0 - v_\mathrm{veh}\right),
\end{equation}
where $k_\mathrm{veh,dec}$ is a design parameter and differs from $k_\mathrm{veh,acc}$. This difference between acceleration and deceleration leads to a more human-like behavior.

	\vspace*{-0.5mm}
	\section{Design of the Decision-Making } \label{sec:cont_des}
	The following presents the simplified simulator consisting of 1) a graphical user interface (GUI), 2) the models of the human motion and the automated vehicle and 3) the real-time implementation of the intention-aware decision-making. Furthermore, a design framework is proposed, which provides an automatic tuning of the parameters using different human models, which is presented in this section.
	
	\vspace*{-0.5mm}
	\subsection{The Simplified Simulator with GUI}
	The testing and verification of the algorithm happen in different stages: 1) pure simulation, 2) human-in-the-loop simplified simulator, 3) human-in-the-loop realistic simulator and 4) real system tests. In this work, the first two stages are used to validate the proposed decision-making algorithm. The benefits of a human-in-the-loop simplified simulator are the following: A realistic human decision-making of the simulated pedestrian can be realized through an user interface (keyboard, joystick) in a test environment. That way sudden changes, un-modeled effects or not easily explainable behavior of the pedestrian can be included in the first testing stages.
	
	The vehicle is modeled as a double integrator along its reference path in the longitudinal direction, since that is a common assumptions, see e.g.~\cite[Chapter 13]{2017_ModernRoboticsMechanics_lynch}.
	\begin{algorithm}
		\caption{The algorithm of the intention-aware decision-making including the functions of the original code}\label{alg:IAA}
		\KwIn{$v_\mathrm{ped}, x_\mathrm{ped},i_\mathrm{ped}, y_\mathrm{veh}$}
		\KwOut{$a_\mathrm{veh}$}
		
		Check:\\
		\hspace{6mm}isPedestrianCrossed\;
		\hspace{6mm}isPedestrianCloseToRoad\;
		\hspace{6mm}isPedestrianGoneThrough\;
		\hspace{6mm}isPedestrianInCollisionArea\;
		\hspace{6mm}isVehicleGoneThrough\;
		
		\While{$\mathrm{not}$ isVehicleGoneThrough $\mathrm{and}$ $\mathrm{not}$ isPedestrianCrossed}{
			
			\uIf{$\mathrm{not}$ canVehSafeCross()}
			{	
				\uIf{isPedestrianInCollisionArea}
				{
					vehStopping()\;
				}
				\uElseIf{isPedestrianGoneThrough}
				{
					vehCrossing()\;
				}
				\uElseIf{isPedestrianCloseToRoad $\mathrm{and}$ $v_\mathrm{ped}>0$}
				{
					vehStopping()\;
				}
				\uElseIf{$v_\mathrm{ped}>v_\mathrm{ped,H}$ $\mathrm{or}$ $i_\mathrm{ped}>i_\mathrm{ped,H}$}
				{
					vehStopping()\;
				}
				\uElseIf{$v_\mathrm{ped,L}<v_\mathrm{ped}<v_\mathrm{ped,H} \; \mathrm{and}$ \hspace*{8mm}$i_\mathrm{ped,L}<i_\mathrm{ped}<i_\mathrm{ped,H}$}
				{
					vehStopping()\;
				}
				\Else
				{
					vehCrossing()\;
				}
			}
			\Else
			{
				vehCrossing()\;
			}
		}

	\end{algorithm}	
	\vspace*{-0.5mm}	
	\subsection{Adaptation of the Pedestrian Models}
	Before optimizing the parameters of the decision-making algorithm, two human models, an SFM and an MDP, are set up with a data set from the state of the art~\cite{2019_TopviewTrajectoriesPedestrian_yang}. Using this open-source data set, a more realistic behavior can be reached and the model fidelity of the simulation is increased. The adaptation of the models includes two scenario:
	\begin{itemize}
		\item[1)] The pedestrian crosses the intersection first forcing the AV to wait.
		\item[2)] The pedestrian waits and lets the AV crosses the intersection first.
	\end{itemize}
	The parameters of the models are obtained through a least-square optimization. Note that for the models of SFM and MDP are tuned with these two scenarios, since most of the common scenarios can be derived from these two.

	\subsection{Parameter Optimization using the Pedestrian Models}
\begin{figure}[t!]
		\centering
		\includegraphics[width=0.85\linewidth]{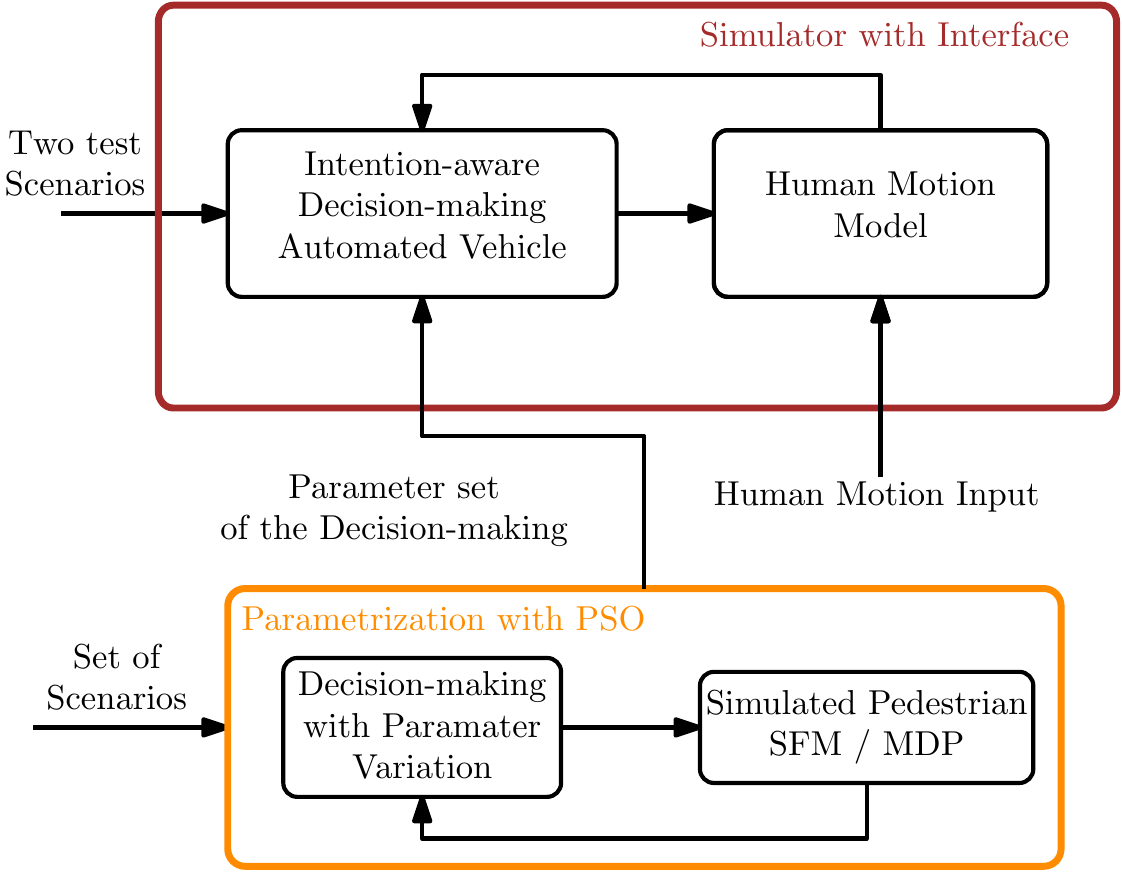}
		\caption{The design framework of the decision-making}
		\label{fig:optim_framework}
	\end{figure}

	On the one hand, the algorithm of the intention-aware decision-making has physically meaningful parameters and simple structure, which is beneficial for testing and the verification of the decision-making algorithm. On the other hand, the considerable number of the parameters makes a manual tuning difficult. Therefore, a design framework is developed, which uses particle swarm optimization (PSO), in order to automatically generate the parameters of the intention-aware decision-making, cf. Fig. \ref{fig:optim_framework}. The result of the PSO is the parameter set of the decision-making algorithm: $i_\mathrm{ped,H}, i_\mathrm{ped,L}, v_\mathrm{ped,H}, v_\mathrm{ped,L}, k_\mathrm{veh,acc}, k_\mathrm{veh,dec},$ and $k_\mathrm{disc}$.	For the optimization, the global objective function 
	\vspace*{-0.5mm}
	\begin{align} \label{eq:f_PSO} \nonumber
		J\play{PSO} = \int_{t_0}^{t_\mathrm{end}}
		k_1 \cdot t &+ k_2 \cdot \left|a^2_\mathrm{max,veh}(t)\right| \\ 
		&- k_3 \cdot \left|d_\mathrm{min}\right| + k_4 \cdot \frac{1}{TTC}(t) \,\mathrm{d}t 	
	\end{align}
	\vspace*{-0.5mm}
	is defined, where $d_\mathrm{min}$ is the minimal distance between AV and VRU during the simulation. The maximal acceleration of the vehicle and the minimum distance between the pedestrian and AV are $a^2_\mathrm{max,veh}$ and $d_\mathrm{min}$, respectively. $k_i$ are subject to design reaching the desired behavior, they are chosen to: 
	$
	k_1 = 1, \; k_2 = 1, \; k_3 = 5, \; k_4 = 0,$ which are easier to choose compared to the parameters of the decision-making.	The parameter $TTC$ is the \textit{time-to-collision} and computed such that
	\begin{equation}
		TTC = \left|\frac{d_\mathrm{veh}}{v_\mathrm{veh} + k_\mathrm{num}}\right|,
	\end{equation}
	where the constant $k_\mathrm{num}>0$ ensures the numerical stability. 

	\vspace*{2mm}
	\section{Results and Discussion} \label{sec:results}

\begin{figure}[t!]
		\centering
		\includegraphics{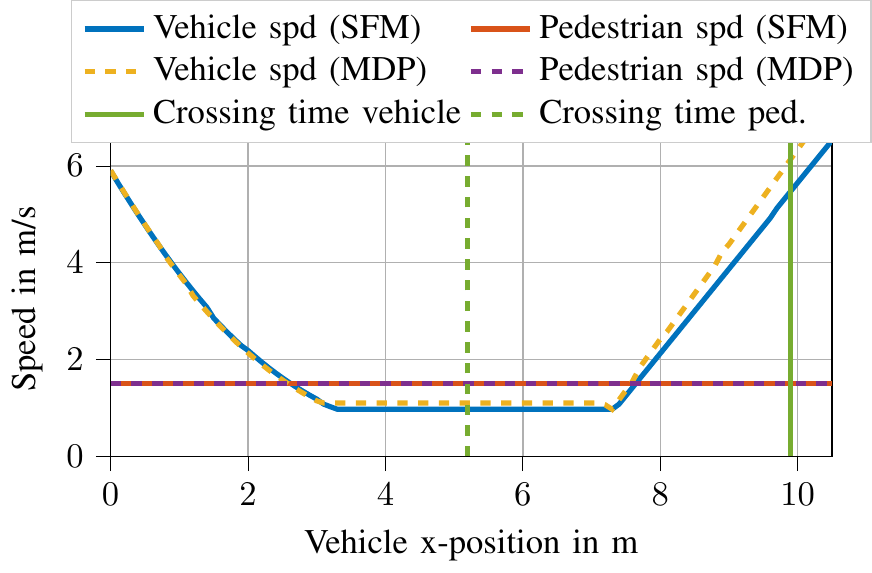}
		\caption{Results of the \textit{normal} test scenario, generated by inputs through the user interface}
		\label{fig:res1_ped_go_first}
	\end{figure}

	The results of the design are compared to each other, since the design yields two parameter sets for the two models applied, leading to different results, cf. Fig.~\ref{fig:optim_framework}. In the first scenario, the pedestrian approaches the street with a constant speed $v_\mathrm{ped} = 1.5 \tfrac{\mathrm{m}}{\mathrm{s}}$ and high intention $i_\mathrm{ped} = 0.55$ to cross the street before the vehicle. It is referred as a normal behavior of the pedestrian. 
Fig.~\ref{fig:res1_ped_go_first} shows the resulting velocity trajectories comparing the two parameter sets designed by the SFM and MDP. It can be seen that the different models have a small impact on the parameters of the designed intention-aware decision-making only.

\begin{figure}[t!]
	\centering
	\includegraphics{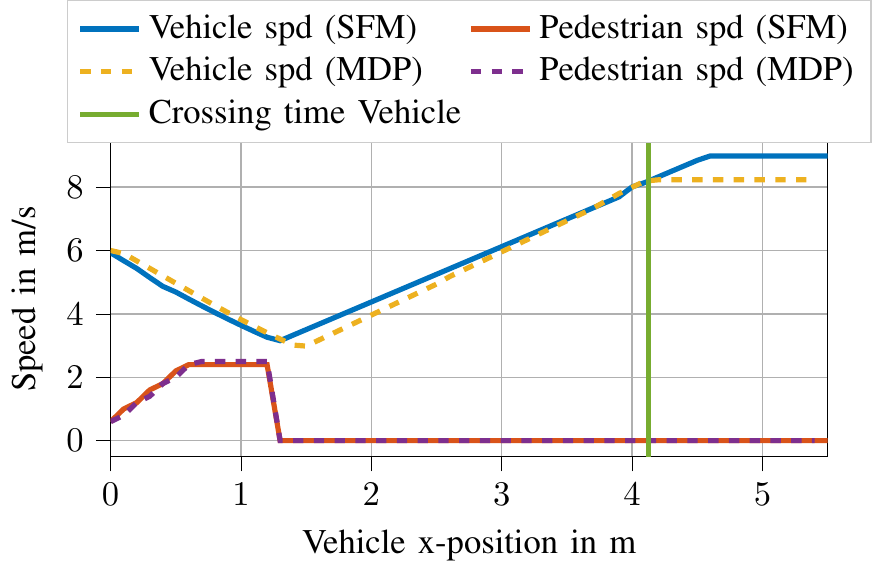}
	\caption{Result of the test scenario with the \textit{unexpected} stopping of the pedestrian without crossing the street}
	\label{fig:res2_ped_go_after}
	\end{figure}

In the second scenario, an unusual case is analyzed, in which the pedestrian approaches the street at a high speed, but they stop, see $t\approx1.2\,$s in Fig.~\ref{fig:res2_ped_go_after} and do not cross the street. In this scenario, the crossing intention of the pedestrian is $i_\mathrm{ped} = 0.2$. Thus, the AV slows down letting through the pedestrian due to their high speed. However, the AV accelerates after the pedestrian stops, since they do not intend to cross the street. As Fig.~\ref{fig:res2_ped_go_after} shows, this unexpected behavior can be also handled by the proposed algorithm indicating the suitability. The two parameter sets obtained from the PSO do not have a significant impact on the resulting velocity trajectories.

	The results indicate that the combination of the physical states (speed and position) and the intention of the pedestrian can resolve un-modeled cases with even a simple structure. 

	\section{Summary and Outlook} \label{sec:summary}
	This paper presented a white-box intention-aware decision-making of an automated vehicle to handle mixed intersection scenarios. One of its most important features is that the procedure is traceable, thus its verification and validation is possible enabling its later real-world use. In order to avoid deadlocks in case of a false estimation, decreasing of the pedestrian's intention is proposed. The further contribution of this paper is a framework, which suits for automated parameter design of the intention-aware decision-making. 
	The results show that there is no significant difference between the overall behaviors designed by the social force model or the Markov decision process and the proposed intention-aware decision-making can resolve unexpected situations as well. In our future work, we plan to apply further concepts using game theory, see \cite{2021_OrdinalPotentialDifferential_varga, 2022_LimitedInformationShared_varga}.
	
	\vspace*{-2mm}
	\section*{Acknowledgment}
	We kindly thank Torsten Schmitt from the version1 GmbH for graphical illustration of the scenario considered in this paper. 
	
	\vspace*{-2mm}
	\bibliographystyle{IEEEtran}


\end{document}